\documentclass{article}
\usepackage{arxiv}
\usepackage{amsmath, amssymb, amsthm}
\usepackage{graphicx}
\usepackage{algorithm}
\usepackage{algpseudocode}
\usepackage[numbers, sort&compress]{natbib}
\usepackage{hyperref}

\title{\textbf{Reinforcement Learning with Anticipation: A Hierarchical Approach for Long-Horizon Tasks}}
\author{Yang Yu\\
National Key Laboratory for Novel Software Technology, Nanjing University, China \\
School of Artificial Intelligence, Nanjing University, China \\
   yuy@nju.edu.cn}
\date{}

\newtheorem{theorem}{Theorem}
\newtheorem{lemma}{Lemma}

\newtheorem{assumption}{Assumption}

\newcommand{\state}{\mathcal{S}}
\newcommand{\action}{\mathcal{A}}

\newcommand{\loss}{\mathcal{L}}
\newcommand{\policy}{\pi}
\newcommand{\anticipator}{\phi}
\newcommand{\critic}{Q}

\newcommand{\replay}{\mathcal{D}}
\newcommand{\goals}{\mathcal{G}}

\begin{document}

\maketitle

\begin{abstract}
Solving long-horizon goal-conditioned tasks remains a significant challenge in reinforcement learning (RL). Hierarchical reinforcement learning (HRL) addresses this by decomposing tasks into more manageable sub-tasks, but the automatic discovery of the hierarchy and the joint training of multi-level policies often suffer from instability and can lack theoretical guarantees. In this paper, we introduce Reinforcement Learning with Anticipation (RLA), a principled and potentially scalable framework designed to address these limitations. The RLA agent learns two synergistic models: a low-level, goal-conditioned policy that learns to reach specified subgoals, and a high-level anticipation model that functions as a planner, proposing intermediate subgoals on the optimal path to a final goal. The key feature of RLA is the training of the anticipation model, which is guided by a principle of value geometric consistency, regularized to prevent degenerate solutions. We present proofs that RLA approaches the globally optimal policy under various conditions, establishing a principled and convergent method for hierarchical planning and execution in long-horizon goal-conditioned tasks.
\end{abstract}

\section{Introduction}

The ability to devise and execute complex plans over long time horizons is a central feature of intelligent behavior. Yet, for artificial agents, this remains a frontier challenge. Reinforcement learning (RL) has demonstrated remarkable success on tasks with clear, immediate feedback, but its effectiveness diminishes dramatically as the gap between actions and their consequences grows. In settings with goal-conditioned tasks, where an agent might perform thousands of actions before receiving a single signal about the task accomplishment, the fundamental problems of exploration and credit assignment can become insurmountable. An agent in this scenario is akin to a navigator trying to find a specific destination in a vast ocean with only a final signal to indicate success, making it nearly impossible to learn which turns were productive and which led astray.

Hierarchical Reinforcement Learning (HRL) has long been proposed as a promising solution to this dilemma \cite{sutton1999between, dietterich2000hierarchical}. The core intuition of HRL is to decompose a daunting, long-horizon task into a more manageable sequence of shorter-term subgoals \cite{pateria2021hierarchical, bartosweep}. This temporal abstraction allows an agent to reason at multiple levels of granularity, focusing on achieving proximal waypoints rather than being overwhelmed by the final objective. However, despite this intuitive appeal, the practical application of HRL has been fraught with persistent difficulties.

First, designing the hierarchy itself is a major hurdle. How are meaningful subgoals defined? If specified by a human designer, the resulting agent is brittle and cannot adapt to new tasks. If they are to be learned automatically, what form should they take, and how can an agent discover them? Second, the process of jointly learning policies at different levels of the hierarchy is notoriously unstable. A high-level policy that selects subgoals has to learn while its low-level counterpart, responsible for executing those subgoals, is also constantly changing. This creates a non-stationary "moving target" problem for the high-level learner, which can severely hinder convergence \cite{levy2017hierarchical, nachum2018data}. Finally, this instability is compounded by the challenge of credit assignment; if a task fails, was it because the high-level policy chose a poor subgoal, or because the low-level policy failed to execute a good one?

Recent advances further highlight practical mechanisms for discovering useful abstractions and subgoals, including temporal state abstraction and waypoint planning \cite{ghosh2022learning}, as well as representation learning that aligns goals with geometric structure in value space \cite{hartikainen2020dynamical,shah2021value,eysenbach2022contrastive}. These developments complement classical HRL by emphasizing shorter-horizon supervision and goal-centric representations, and they motivate our anticipatory subgoal generation driven by value-geometry.

This paper introduces a new framework, Reinforcement Learning with Anticipation (RLA), designed to overcome these fundamental challenges through a principled approach to hierarchical control. Our approach decomposes the agent into two synergistic components that can be understood intuitively as a high-level navigator and a low-level motor controller. The low-level policy, our motor controller, is a standard goal-conditioned agent that learns the simple skill of ``how to get to a nearby location". The high-level component, which we call the anticipation model, acts as the navigator, addressing the strategic question: ``Given my current location and my ultimate destination, what is an optimal intermediate waypoint to aim for?"

The key innovation of RLA lies in how this high-level navigator learns. Instead of relying on the sparse and delayed reward from the overall task, the anticipation model learns by inspecting the agent's value function. We can think of the learned value function as a map of the environment that estimates the ``travel time" between any two points. A smart waypoint from a starting point to a destination is one that lies on the most direct route. Such a waypoint has a unique geometric property: the travel time from the start to the destination is precisely the sum of the time from the start to the waypoint and the time from the waypoint to the destination. Any other point would represent a detour.

The RLA anticipation model is trained to find such waypoints that satisfy this condition of optimality. This design can sidestep the classic HRL challenges. The subgoals are the states set by the anticipation model, and the hierarchy is simply constructed recursively, which is discovered automatically but not pre-defined. Most importantly, the learning objective for the high-level navigator is decoupled from the direct outcome of the low-level policy's actions, instead deriving its signal from the stable structure of the value function. This principled design allows us to prove that the agent (approximately) converges to a globally optimal policy under some conditions. 

This paper proceeds by presenting the problem setting, our RLA framework, the theoretical analysis, related work, and a concluding discussion.
 
\section{Preliminaries and Problem Setting}

We formalize the problem of learning to solve long-horizon tasks using the framework of a Goal-Conditioned Markov Decision Process (GMDP). A GMDP is an extension of a standard MDP where the agent's behavior is conditioned on achieving a specific goal, which is fit for multi-task scenarios.

\subsection{Goal-conditioned Markov decision processes}
A goal-conditioned Markov decision process (GMDP) is defined by the tuple $(\mathcal{S}, \mathcal{B},\mathcal{A}, \mathcal{G}, P, R, H)$, where:
\begin{itemize}
    \item $\mathcal{S}$ is the set of states.
    \item $\mathcal{B}$ is the set of initial states.
    \item $\mathcal{A}$ is the set of actions.
    \item $\mathcal{G}$ is the set of possible goals. For the problems we consider, the goal is to reach a particular state, so we can assume the goal space is equivalent to the state space, $\mathcal{G} \equiv \mathcal{S}$.
    \item $P: \mathcal{S} \times \mathcal{A} \to \Delta(\mathcal{S})$ is the state transition function, where $P(s'|s,a)$ gives the probability of transitioning to state $s'$ after taking action $a$ in state $s$.
    \item $R: \mathcal{S} \times \mathcal{A} \times \mathcal{G} \to \mathbb{R}$ is the goal-conditioned reward function.
    \item $H$ is the maximum horizon.
\end{itemize}

The agent's objective is to learn a policy that can achieve any goal $g \in \mathcal{G}$ from any state $s \in \mathcal{S}$. A task is thus defined by a start-goal pair $(s_0, g)$.

\subsection{Objective and value functions}
The agent interacts with the environment using a goal-conditioned policy, $\policy(a|s, g)$, which specifies a probability distribution over actions given the current state $s$ and the desired goal $g$. The objective of the agent is to learn an optimal policy $\policy^*$ that maximizes the expected cumulative reward for any given task. We use an undiscounted formulation ($\gamma=1$) because the central value decomposition theory relies on it.
\begin{equation}
    \policy^* = \arg\max_{\policy} \mathbb{E}_{\tau \sim \policy(\cdot|s,g)} \left[ \sum_{t=0}^{H-1} R(s_t, a_t, g) \right] \quad \forall s \in \state, g \in \goals
\end{equation}
where $\tau = (s_0, a_0, \dots, s_{H-1}, a_{H-1}, s_H=g)$ is a trajectory of length $H$. Crucially, because each task is episodic and terminates upon reaching the goal, the horizon $H$ is finite, ensuring that the undiscounted sum of rewards is well-defined and avoids the infinite value problem common in continuing tasks.

Associated with any policy $\policy$ are the goal-conditioned state-value function $V^\policy(s, g)$ and action-value function $Q^\policy(s, a, g)$, which represent the expected return from state $s$ (or state-action pair $(s,a)$) when pursuing goal $g$:
\begin{align}
    V^\policy(s, g) &= \mathbb{E}_{\tau \sim \policy(\cdot|s,g)} \left[ \sum_{t=0}^{H-1} R(s_t, a_t, g) \Big| s_0 = s \right] \\
    Q^\policy(s, a, g) &= \mathbb{E}_{\tau \sim \policy(\cdot|s,g)} \left[ \sum_{t=0}^{H-1} R(s_t, a_t, g) \Big| s_0 = s, a_0 = a \right]
\end{align}

The optimal value functions, $V^*$ and $Q^*$, satisfy the Bellman optimality equations for this undiscounted, episodic setting:
\begin{align}
    \critic^*(s, a, g) &= R(s, a, g) + \mathbb{E}_{s' \sim P(\cdot|s,a)} \left[ V^*(s', g) \right] \\
    V^*(s, g) &= \max_{a} \critic^*(s, a, g)
\end{align}

\subsection{The challenge of long-horizon sparse-reward tasks}
The primary challenge we address arises from the structure of the reward function in typical goal-directed tasks. To encourage the agent to reach the goal state $g$ as efficiently as possible, a sparse reward function is commonly used:
\begin{equation}
    R(s_t, a_t, g) = 
    \begin{cases}
        0 & \text{if } s_{t+1} = g \\
        -1 & \text{if } s_{t+1} \neq g
    \end{cases}
\end{equation}
Under this reward structure, maximizing the expected return is equivalent to minimizing the number of steps taken to reach the goal. The optimal value function $V^*(s, g)$ thus corresponds to the negative shortest path distance from $s$ to $g$.

When the distance between a starting state and a goal is large, the agent must execute a long sequence of actions before receiving the non-penalizing reward of 0. This delay makes it incredibly difficult to perform credit assignment and determine which actions contributed to the eventual success, posing a significant obstacle to learning. Our work aims to provide a hierarchical solution that can effectively solve such tasks by discovering intermediate waypoints to bridge these long temporal gaps.

\section{Framework of Reinforcement Learning with Anticipation }

To address the challenges of long-horizon, sparse-reward tasks, we propose the Reinforcement Learning with Anticipation (RLA) framework. The core idea is to decompose the monolithic problem of ``get from start to finish" into a repeating, simpler problem: ``from my current location, what is the next optimal waypoint to aim for on the path to the finish?" This decomposition is managed by a two-level hierarchy.

\subsection{Anticipation model}

The heart of the RLA framework is the anticipation model, $\anticipator_\psi(s_0, s_g)$, where $\psi$ represents the parameters of a neural network. This model is a function that takes the current state $s_0$ and the final desired goal state $s_g$ as input, and outputs a proposed intermediate subgoal, $\hat{s} = \anticipator_\psi(s_0, s_g)$.

The purpose of this model is to identify a waypoint that lies on an optimal shortest path. This objective is rooted in the triangle inequality: for any three points, the shortest path between two of them is never longer than the path that goes through the third. For a point $\hat{s}$ to lie on the shortest path between $s_0$ and $s_g$, this inequality must become an equality: the distance from $s_0$ to $s_g$ must be exactly the sum of the distance from $s_0$ to $\hat{s}$ and the distance from $\hat{s}$ to $s_g$.

Recalling from our problem setting that the optimal value function $V^*(s, g)$ in a shortest-path problem represents the negative distance, $-d(s, g)$, we can translate this geometric condition into the language of value functions. The anticipation model is trained to find a subgoal $\hat{s}$ that satisfies the following condition of optimality:
\begin{equation}
    V^*(s_0, s_g) = V^*(s_0, \hat{s}) + V^*(\hat{s}, s_g)
\end{equation}
This provides a principled, mathematical objective for what constitutes a good subgoal, forming the foundation of our training procedure.

\subsection{Inference procedure: hierarchical planning and acting}

At inference time, when the agent must solve a task, it uses its two modules in a closed loop to generate behavior. Given a starting state $s_0$ and a final goal $s_g$, the agent executes an iterative procedure. 

First, the agent consults its high-level anticipation model by feeding its current state $s_t$ and the final goal $s_g$ into the anticipation model. This generates an intermediate subgoal, $\hat{s} = \anticipator_\psi(s_t, s_g)$. The anticipation model can be recursively invoked, $\hat{s} = \anticipator_\psi(s_t, \hat{s})$, for $J$ times, to generate a subgoal close to the current state. This parameter $J$ allows control over the proximity of the generated subgoal.

Following this, the agent passes control to its low-level motor controller. For a fixed block of $K$ time steps, the low-level policy $\policy_\theta(s, \hat{s})$ takes actions in the environment with the sole objective of reaching the subgoal $\hat{s}$. After the $K$ steps are completed, the agent finds itself at a new state, $s_{t+K}$. 

The agent then repeats the entire process: it re-evaluates its position, anticipates a new waypoint on the path from its new state to the final goal, and acts to reach it. This plan-act cycle continues until the agent's state is within a threshold distance of the final goal $s_g$. Lines 8 to 18 in Algorithm \ref{alg:rla} describe the inference procedure.

\subsection{Training procedure: jointly learning to plan and act}

The model components of the RLA agent are trained jointly from experience collected in a replay buffer $\replay$. However, joint training presents a chicken-and-egg problem: the anticipation model learns from the critic's value function, but the critic needs good data from a competent policy to learn an accurate value function. A randomly initialized critic provides a poor training signal, which can lead to a poor anticipation model, which in turn leads to poor exploration and a failure to improve the critic.

To address this, we introduce a two-phase training schedule. The first phase is a warm-up period where only the low-level actor-critic module is trained. During this phase, subgoals are generated randomly or heuristically, allowing the low-level policy to learn basic motor skills and populate the replay buffer with diverse, short-horizon experiences. This ensures the critic develops a reasonably accurate value estimate for nearby states before it is used to train the high-level planner.

After the warm-up phase, the full joint training begins. The process consists of two parallel updates.

The first part of the training continues to focus on the low-level policy using a standard off-policy actor-critic algorithm (e.g., DDPG \cite{lillicrap2015continuous}) with Hindsight Experience Replay (HER) \cite{andrychowicz2017hindsight}. HER is crucial for providing dense learning signals, allowing the actor and critic to learn efficiently from every interaction by relabeling failed trajectories as successful attempts at goals that were actually reached.

The second part of the training develops the high-level anticipation model, $\anticipator_\psi$, using a regularized loss function derived from the critic's value estimates. For a sampled start-end pair $(s_i, s_j)$, the model generates a subgoal $\hat{s} = \anticipator_\psi(s_i, s_j)$. The loss function is designed to enforce two properties: that the subgoal lies on the shortest path and that it is a non-trivial, intermediate waypoint. The total loss is:
\begin{equation}
\label{eq:anticipation_loss}
\loss_\psi(s_i, s_j) = \loss_{detour} + \lambda (\loss_{prog} + \loss_{non\_trivial})
\end{equation}
where:
\begin{itemize}
    \item $\loss_{detour} = \text{ReLU}\left( V(s_i, s_j) - V(s_i, \hat{s}) - V(\hat{s}, s_j) \right)$ penalizes subgoals that violate the shortest-path equality, where $V$ is estimated by the target critic.
    \item $\loss_{prog} = \text{ReLU}(c_{prog} + V(s_i, \hat{s}))$ ensures the subgoal represents meaningful progress. Since values are non-positive, this term penalizes subgoals that are closer than a margin $c_{prog}$ to the start state $s_i$, preventing the degenerate solution $\hat{s} \approx s_i$.
    \item $\loss_{non\_trivial} = \text{ReLU}(c_{non\_trivial} + V(\hat{s}, s_j))$ ensures the subgoal is not the final goal itself by penalizing subgoals that are too close to $s_j$.
\end{itemize}
The hyperparameter $\lambda$ balances the primary geometric objective with the regularization. This regularized loss provides a robust signal, guiding $\anticipator_\psi$ to produce useful, non-degenerate waypoints. Lines 20 to 33 in Algorithm \ref{alg:rla} describe the training updates.

\begin{algorithm}[H]
\caption{Reinforcement Learning with Anticipation (RLA)}
\label{alg:rla}
\begin{algorithmic}[1]
\State Initialize actor $\policy_\theta$, critic $\critic_\omega$, anticipation model $\anticipator_\psi$, and replay buffer $\replay$.
\State Initialize target networks: $\theta' \leftarrow \theta$, $\omega' \leftarrow \omega$.
\State Initialize hyperparameters: warm-up steps $N_{warmup}$, regularization weight $\lambda$, margins $c_{prog}, c_{non\_trivial}$.

\For{each episode}
    \State Sample a start state $s_0$ and a final goal $s_g$. Let $s \leftarrow s_0$.
    \State $\tau=\emptyset$.
    \State \Comment{Inference with exploration}
    \For{$i = 0, 1, 2, \dots, H/K - 1$}
        \If{episode $>$ $N_{warmup}$}
            \State Let $\hat{s}=s_g$, generate subgoal $\hat{s} = \anticipator_\psi(s, \hat{s})$ for $J$ times.
        \Else
            \State Generate a random subgoal $\hat{s}$ for exploration.
        \EndIf
        \State Execute policy $\policy_\theta(s, \hat{s})$ for $K$ steps, collecting transitions.
        \State Record trajectory $\tau = \tau \cup \{(s_k, a_k, r_k, s'_{k}, \hat{s})\}_{k=1}^K$.
        \State $s \leftarrow s'_{K-1}$ \Comment{Update current state}
    \EndFor
    \State Store $\tau$ in replay buffer $\replay$.
        
    \State \Comment{Training updates}
    \For{$N$ update steps}
        \State Sample a minibatch of transitions and relabel with HER, creating $\replay'$.
        \State \Comment{Update low-level actor-critic}
        \State Update critic $\critic_\omega$ by minimizing Bellman error loss on samples from $\replay'$.
        \State Update actor $\policy_\theta$ using the deterministic policy gradient on samples from $\replay'$.
        
        \If{episode $>$ $N_{warmup}$}
            \State \Comment{Update anticipation model}
            \State Sample a minibatch of start-end state pairs $(s_i, s_j)$ from $\replay'$.
            \State Generate subgoals: $\hat{s} \leftarrow \anticipator_\psi(s_i, s_j)$.
            \State Compute values using target critic: $V(s,g) = \max_a \critic_{\omega'}(s,a,g)$.
            \State Compute regularized anticipation loss $\loss_\psi$ from Eq. \eqref{eq:anticipation_loss}.
            \State Update anticipation model $\anticipator_\psi$ by descending the gradient $\nabla_\psi \loss_\psi$.
        \EndIf
    \EndFor
    \State \Comment{Update target networks}
    \State $\theta' \leftarrow \tau\theta + (1-\tau)\theta'$
    \State $\omega' \leftarrow \tau\omega + (1-\tau)\omega'$
\EndFor
\end{algorithmic}
\end{algorithm}

\section{Theoretical Analysis}

In this section, we provide a formal analysis of the RLA framework. We begin by proving that RLA converges to a globally optimal policy in an idealized deterministic setting. We then extend this analysis to the more practical case involving bounded approximation and stochastic environments.

\subsection{Idealized deterministic setting}

We firstly consider the following set of assumptions, which are common in classical theoretical analysis of reinforcement learning algorithms.

\begin{assumption}[Idealized Conditions]
\label{ass:ideal}
We assume the learning environment and process satisfy the following four conditions:
\begin{enumerate}
    \item \textbf{Tabular and finite spaces:} The state space $\state$ and action space $\action$ are finite.
    \item \textbf{Deterministic dynamics:} The environment's transition function $P(s'|s,a)$ is deterministic. For any state-action pair $(s,a)$, there is only one possible next state $s'$.
    \item \textbf{Shortest-path reward structure:} The reward is -1 for every step, except for a reward of 0 for transitioning to the goal state. This makes maximizing the cumulative reward equivalent to finding the shortest path. Consequently, the optimal value function $V^*(s, g)$ is the negative of the shortest-path distance between states $s$ and $g$, denoted as $-d(s, g)$.
    \item \textbf{Sufficient exploration:} As we introduce randomization in the generation of actions and sub-goals, we can assume that the training process guarantees that every state-action pair is visited an infinite number of times, and every state is used as a goal in training.
\end{enumerate}
\end{assumption}

These assumptions, while simplifying the problem, provide a crucial foundation for theoretical guarantees. The tabular and deterministic setting is a standard starting point for RL convergence proofs and is satisfied in many classic control problems like grid worlds or discretized robotic manipulation tasks. The shortest-path reward structure is the natural formulation for goal-reaching tasks. Finally, the sufficient exploration assumption is a theoretical requirement for tabular methods to ensure they learn about the entire environment, a condition that practical exploration strategies aim to approximate.

With these assumptions in place, we can state our main theoretical result.

\begin{theorem}[Global Optimality of RLA]
\label{thm:main1}
Under Assumption \ref{ass:ideal}, the RLA framework converges to a globally optimal policy. The converged anticipation model, $\anticipator_\psi^*$, produces subgoals that lie on a shortest path to the final goal; the converged low-level policy, $\policy_\theta^*$, executes the sequence of actions that optimally reaches those subgoals. Consequently, the agent's behavior traces a shortest path from any start state to any goal state.
\end{theorem}

To prove this theorem, we first establish two lemmas concerning the convergence of the low-level module and the geometric properties of the learned value function.

\begin{lemma}[Convergence of the Low-Level Module]
\label{lemma:critic}
Under Assumption \ref{ass:ideal}, the low-level actor-critic module converges to the optimal critic $\critic^*$ and policy $\policy^*$. The value function derived from the converged critic, $V_\omega(s, g) = \max_a \critic_\omega(s, a, g)$, becomes equivalent to the true optimal value function, $V^*(s, g)$, which is the negative shortest-path distance $-d(s, g)$.
\end{lemma}

\begin{proof}
The proof rests on demonstrating that our training procedure for the low-level module satisfies the conditions for the convergence of tabular Q-learning for every possible goal in the goal space $\mathcal{G}$. For a fixed goal $g$, the problem reduces to a standard MDP. The classic proof for the convergence of tabular Q-learning states that the estimated Q-values will converge to the optimal Q-values if every state-action pair is visited and updated an infinite number of times \cite{watkins1992q}.

Under Assumption \ref{ass:ideal}(4), the agent's exploration strategy (including the warm-up phase) ensures every state-action pair is executed infinitely often. When these interactions occur, the resulting transitions are stored in the replay buffer. The HER algorithm then ensures that for any state-action pair $(s_i, a_i)$ and any arbitrary goal state $g_j$ that appears later in the same trajectory, a hindsight training example is generated. The sufficient exploration assumption guarantees that such trajectories will be generated infinitely often for all $(s_i, a_i, g_j)$ combinations. Therefore, the conditions for the convergence of tabular Q-learning are met for every goal-specific subproblem.

Consequently, the critic $\critic_\omega$ converges to the true optimal goal-conditioned action-value function $\critic^*(s, a, g)$, the value function $V_\omega(s, g)$ converges to $V^*(s, g) = -d(s, g)$, and the actor $\policy_\theta$ converges to the optimal policy $\policy^*(s, g)$.
\end{proof}

\begin{lemma}[Value Triangle Inequality]
\label{lemma:triangle}
Let $V^*(s, g) = -d(s, g)$ be the optimal value function under the shortest-path reward structure. For any three states $s_i, s_j, z \in \state$, the value function must satisfy the inequality $V^*(s_i, s_j) \ge V^*(s_i, z) + V^*(z, s_j)$. Equality holds if and only if the state $z$ lies on a shortest path from $s_i$ to $s_j$.
\end{lemma}

\begin{proof}
This property is a direct consequence of the triangle inequality for shortest-path distances, $d(s_i, s_j) \le d(s_i, z) + d(z, s_j)$. Substituting our definition of the optimal value function, $d(s, g) = -V^*(s, g)$, and multiplying by -1 reverses the inequality, yielding $V^*(s_i, s_j) \ge V^*(s_i, z) + V^*(z, s_j)$. The condition for equality remains unchanged.
\end{proof}

\begin{proof}[Proof of Theorem \ref{thm:main1}]
The proof proceeds by analyzing the convergence of the high-level anticipation model, given the established convergence of the low-level module from Lemma \ref{lemma:critic}.

From Lemma \ref{lemma:critic}, under our assumptions, the low-level training procedure ensures that the critic's value estimate $V_\omega(s,g)$ converges to the true optimal value function $V^*(s,g) = -d(s,g)$. As training progresses, the target network used in the anticipation loss will also converge to $V^*$.

The anticipation model $\anticipator_\psi$ is trained by gradient descent to minimize the regularized loss function $\loss_\psi$ from Eq. \eqref{eq:anticipation_loss}. Let us analyze this loss function once the critic has converged. Let $\hat{s} = \anticipator_\psi(s_i, s_j)$ be the subgoal proposed by the model for a given start-end pair $(s_i, s_j)$. The loss is:
\begin{align*}
\loss_\psi(s_i, s_j) = & \text{ReLU}\left( V^*(s_i, s_j) - V^*(s_i, \hat{s}) - V^*(\hat{s}, s_j) \right) \\&+ \lambda \left( \text{ReLU}(c_{prog} + V^*(s_i, \hat{s})) + \text{ReLU}(c_{non\_trivial} + V^*(\hat{s}, s_j)) \right)
\end{align*}
The training process seeks parameters $\psi$ that achieve the global minimum of this loss function. Since each `ReLU` term is non-negative and $\lambda > 0$, the entire loss $\loss_\psi$ is non-negative. Therefore, the global minimum of $\loss_\psi$ is 0.

The loss $\loss_\psi$ achieves its minimum of 0 if and only if all of its constituent non-negative terms are simultaneously equal to 0. This gives us three conditions that must be satisfied by the subgoal $\hat{s}$ generated by the converged model $\anticipator_\psi^*$:

\begin{enumerate}
    \item \textbf{Shortest Path Condition:} The detour loss must be zero.
    \begin{equation*}
    \text{ReLU}\left( V^*(s_i, s_j) - V^*(s_i, \hat{s}) - V^*(\hat{s}, s_j) \right) = 0
    \end{equation*}
    This implies $V^*(s_i, s_j) - V^*(s_i, \hat{s}) - V^*(\hat{s}, s_j) \le 0$. However, from the Value Triangle Inequality (Lemma \ref{lemma:triangle}), we know that $V^*(s_i, s_j) \ge V^*(s_i, \hat{s}) + V^*(\hat{s}, s_j)$ always holds. For both inequalities to be true, they must be equal:
    \begin{equation*}
    V^*(s_i, s_j) = V^*(s_i, \hat{s}) + V^*(\hat{s}, s_j)
    \end{equation*}
    By Lemma \ref{lemma:triangle}, this equality holds if and only if the subgoal $\hat{s}$ lies on a shortest path from $s_i$ to $s_j$.

    \item \textbf{Progress Condition:} The progress regularization loss must be zero.
    \begin{equation*}
    \text{ReLU}(c_{prog} + V^*(s_i, \hat{s})) = 0
    \end{equation*}
    This implies $c_{prog} + V^*(s_i, \hat{s}) \le 0$. Substituting $V^*(s, g) = -d(s, g)$, we get $c_{prog} - d(s_i, \hat{s}) \le 0$, which is equivalent to $d(s_i, \hat{s}) \ge c_{prog}$. By choosing a positive margin $c_{prog} > 0$ (e.g., $c_{prog}=1$, the cost of a single step), this condition ensures that the subgoal $\hat{s}$ is a non-zero distance away from the start state $s_i$. This explicitly prevents the degenerate solution $\hat{s} = s_i$.

    \item \textbf{Non-Triviality Condition:} The non-triviality regularization loss must be zero.
    \begin{equation*}
    \text{ReLU}(c_{non\_trivial} + V^*(\hat{s}, s_j)) = 0
    \end{equation*}
    This implies $c_{non\_trivial} + V^*(\hat{s}, s_j) \le 0$. Substituting the value-distance relationship gives $c_{non\_trivial} - d(\hat{s}, s_j) \le 0$, or $d(\hat{s}, s_j) \ge c_{non\_trivial}$. By choosing $c_{non\_trivial} > 0$, this condition ensures the subgoal $\hat{s}$ is also a non-zero distance away from the end state $s_j$, preventing the degenerate solution $\hat{s} = s_j$.
\end{enumerate}

Combining these three conditions, a converged anticipation model $\anticipator_\psi^*$ that minimizes the loss to its global optimum must produce a subgoal $\hat{s}$ that is a \textbf{non-trivial waypoint on a shortest path} to the goal.

Since the low-level policy $\policy_\theta^*$ has also converged (by Lemma \ref{lemma:critic}) to be an optimal policy for reaching any given subgoal, the complete RLA agent's behavior is as follows: at any state, it generates a valid, intermediate subgoal on the globally optimal path to the final destination and then executes the optimal policy to reach that subgoal. By induction, the sequence of actions taken by the agent traces a globally optimal, shortest path to the final goal. This completes the proof.
\end{proof}

\subsection{Bounded error assumptions}

Instead of assuming perfect convergence, we adopt a set of bounded error assumptions. This is standard practice in the analysis of large-scale reinforcement learning algorithms that use function approximation.

\begin{assumption}[Bounded Errors Assumptions]
\label{ass:bounded_error}
We assume the learning process is effective enough to bound the errors in each component:
\begin{enumerate}
    \item \textbf{Bounded value function error:} The low-level critic is trained such that its converged value estimate $V_\omega$ is uniformly close to the true optimal value function $V^*$. There exists a small constant $\epsilon_V \ge 0$ such that for all states $s$ and goals $g$:
    \begin{equation}
        |V_\omega(s, g) - V^*(s, g)| \le \epsilon_V
    \end{equation}
    This error $\epsilon_V$ encapsulates inaccuracies from function approximation, finite sampling, and imperfect optimization.

    \item \textbf{Bounded anticipation loss:} The high-level anticipation model is trained to minimize its loss function. We assume the training procedure finds a model $\anticipator_\psi$ that achieves a loss value no greater than a small constant $\epsilon_\psi \ge 0$ for its primary geometric objective.
    \begin{equation}
        \max(0, V_\omega(s, g) - V_\omega(s, \hat{s}) - V_\omega(\hat{s}, g) ) \le \epsilon_\psi
    \end{equation}
    where $\hat{s} = \anticipator_\psi(s, g)$.

    \item \textbf{Bounded policy sub-optimality:} The low-level policy is imperfect. When tasked with reaching a subgoal $\hat{s}$ from a state $s$, it incurs a total cost (negative cumulative reward) of $C_\theta(s, \hat{s})$. We assume this cost is bounded by the optimal cost (the true shortest path distance $d(s, \hat{s})$) plus a sub-optimality gap $\epsilon_\pi \ge 0$.
    \begin{equation}
        C_\theta(s, \hat{s}) \le d(s, \hat{s}) + \epsilon_\pi
    \end{equation}
\end{enumerate}
\end{assumption}

These assumptions are more practical than their idealized counterparts. They acknowledge that neural networks will never be perfect and instead quantify their quality, allowing us to analyze how these imperfections propagate through the hierarchical system.
With these assumptions, we can prove that the overall performance of the RLA agent is gracefully bounded.

\begin{theorem}[Bounded Sub-optimality of RLA]
\label{thm:main_error}
Under the bounded error conditions in Assumption \ref{ass:bounded_error}, the total cost $C_{RLA}(s_0, s_g)$ for the RLA agent to travel from a start state $s_0$ to a goal state $s_g$ over $M$ high-level steps is bounded by:
\begin{equation}
    C_{RLA}(s_0, s_g) \le d(s_0, s_g) + M \cdot (\epsilon_\pi + 3\epsilon_V + \epsilon_\psi)
\end{equation}
This means the total sub-optimality of the agent's path, $C_{RLA} - d(s_0, s_g)$, grows linearly with the number of high-level steps, and the error per step is a sum of the fundamental component errors.
\end{theorem}

To prove this theorem, we first establish a key lemma that bounds the detour created by each anticipated subgoal as a function of the underlying learning errors.

\begin{lemma}[Bounded Detour of Anticipated Subgoals]
\label{lemma:detour}
Given Assumption \ref{ass:bounded_error}, the subgoal $\hat{s} = \anticipator_\psi(s, g)$ generated by the anticipation model creates a one-step path whose true detour cost is bounded. Specifically, the amount by which the path through the subgoal exceeds the direct path is bounded as:
\begin{equation}
    d(s, \hat{s}) + d(\hat{s}, g) - d(s, g) \le 3\epsilon_V + \epsilon_\psi
\end{equation}
\end{lemma}

\begin{proof}
The proof proceeds by relating the true detour cost, which is defined by the true distance function $d$, to the anticipation loss, which is defined by the learned value function $V_\omega$.

The true detour cost is $D = d(s, \hat{s}) + d(\hat{s}, g) - d(s, g)$.
Recalling that $d(s, g) = -V^*(s, g)$, we can write the detour cost in terms of the optimal value function:
\begin{equation}
    D = -V^*(s, \hat{s}) - V^*(\hat{s}, g) + V^*(s, g)
\end{equation}
We now use the bounded value function error from Assumption \ref{ass:bounded_error}(1), which states $V^*(x, y) \le V_\omega(x, y) + \epsilon_V$ and $V^*(x, y) \ge V_\omega(x, y) - \epsilon_V$. To get an upper bound on $D$, we substitute the bounds for each term accordingly:
\begin{align*}
    D & \le -(V_\omega(s, \hat{s}) - \epsilon_V) - (V_\omega(\hat{s}, g) - \epsilon_V) + (V_\omega(s, g) + \epsilon_V) \\
    &  = V_\omega(s, g) - V_\omega(s, \hat{s}) - V_\omega(\hat{s}, g) + 3\epsilon_V
\end{align*}
From Assumption \ref{ass:bounded_error}(2), we know that the anticipation loss is bounded:
\begin{equation*}
    V_\omega(s, g) - V_\omega(s, \hat{s}) - V_\omega(\hat{s}, g) \le \epsilon_\psi
\end{equation*}
Substituting this into our inequality for $D$:
\begin{equation}
    D \le \epsilon_\psi + 3\epsilon_V
\end{equation}
This completes the proof.
\end{proof}

\begin{proof}[Proof of Theorem \ref{thm:main_error}]
Let the sequence of states chosen by the high-level planner be $s_0, s_1, s_2, \dots, s_M$, where $s_0$ is the start state and $s_M$ is the final goal $s_g$. At each step $k \in \{0, \dots, M-1\}$, the agent is at state $s_k$ and the anticipation model proposes the next subgoal $s_{k+1} = \anticipator_\psi(s_k, s_g)$. For this analysis, we assume the low-level policy successfully reaches the subgoal.

The total cost incurred by the RLA agent is the sum of the costs of executing the low-level policy between consecutive subgoals:
\begin{equation}
    C_{RLA}(s_0, s_g) = \sum_{k=0}^{M-1} C_\theta(s_k, s_{k+1})
\end{equation}
Using the bounded policy sub-optimality from Assumption \ref{ass:bounded_error}(3), we can bound the cost of each segment:
\begin{equation}
    C_{RLA}(s_0, s_g) \le \sum_{k=0}^{M-1} (d(s_k, s_{k+1}) + \epsilon_\pi) = \left(\sum_{k=0}^{M-1} d(s_k, s_{k+1})\right) + M \cdot \epsilon_\pi
\end{equation}
The core of the proof is to bound the total path length of the subgoals, $\sum d(s_k, s_{k+1})$. We use our key lemma on the bounded detour cost. From Lemma \ref{lemma:detour}, for each step $k$, we have:
\begin{equation}
    d(s_k, s_{k+1}) + d(s_{k+1}, s_g) - d(s_k, s_g) \le 3\epsilon_V + \epsilon_\psi
\end{equation}
Rearranging gives:
\begin{equation}
    d(s_k, s_{k+1}) \le d(s_k, s_g) - d(s_{k+1}, s_g) + 3\epsilon_V + \epsilon_\psi
\end{equation}
Now, we can sum this inequality over the entire sequence of $M$ steps:
\begin{equation}
    \sum_{k=0}^{M-1} d(s_k, s_{k+1}) \le \sum_{k=0}^{M-1} \left( d(s_k, s_g) - d(s_{k+1}, s_g) \right) + M \cdot (3\epsilon_V + \epsilon_\psi)
\end{equation}
The first term on the right-hand side is a telescoping sum:
\begin{equation}
    (d(s_0, s_g) - d(s_1, s_g)) + (d(s_1, s_g) - d(s_2, s_g)) + \dots + (d(s_{M-1}, s_g) - d(s_M, s_g)) = d(s_0, s_g) - d(s_M, s_g)
\end{equation}
Since the final state is the goal, $s_M = s_g$, the last term is $d(s_g, s_g) = 0$. The telescoping sum simplifies to $d(s_0, s_g)$.
Substituting this back, we get a bound on the total path length:
\begin{equation}
    \sum_{k=0}^{M-1} d(s_k, s_{k+1}) \le d(s_0, s_g) + M \cdot (3\epsilon_V + \epsilon_\psi)
\end{equation}
Finally, we substitute this result into our bound for the total cost $C_{RLA}$:
\begin{equation}
    C_{RLA}(s_0, s_g) \le \left( d(s_0, s_g) + M \cdot (3\epsilon_V + \epsilon_\psi) \right) + M \cdot \epsilon_\pi
\end{equation}
\begin{equation}
    C_{RLA}(s_0, s_g) \le d(s_0, s_g) + M \cdot (\epsilon_\pi + 3\epsilon_V + \epsilon_\psi)
\end{equation}
This completes the proof.
\end{proof}

\subsection{Stochastic environments}

We now extend our analysis to stochastic environments. This is a more realistic setting where an agent's actions have probabilistic outcomes. The key challenge is that the low-level policy can no longer guarantee reaching a specific subgoal state. Instead, it can only guarantee reaching a distribution of states in the vicinity of the subgoal. 

To address this, we refine our assumptions. The most critical change is acknowledging that some states may not be reachable from others. We introduce the standard assumption of a communicating MDP, which ensures the entire state space is connected.

\begin{assumption}[Bounded Errors in a Communicating Stochastic MDP]
\label{ass:stochastic_error}
We assume the learning environment is a finite goal-conditioned Markov decision process (GMDP) and that the learning process is effective enough to bound the errors in its components:
\begin{enumerate}
    \item \textbf{Communicating GMDP:} The GMDP is communicating, meaning that for any two states $s_i, s_j \in \mathcal{S}$, there exists a policy that can reach $s_j$ from $s_i$ with a non-zero probability. This ensures that no part of the state space is permanently inaccessible.
    \item \textbf{Expected cost objective:} The agent receives a reward of -1 for every step. The objective is to find a policy that reaches the goal state while minimizing the expected total number of steps. The optimal value function, $V^*(s, g)$, represents the negative of the minimum expected number of steps (expected hitting time) to get from state $s$ to goal $g$.
    \item \textbf{Bounded value function error:} The low-level critic converges to a value function $V_\omega$ that is uniformly close to the true optimal expected value function $V^*$. There exists a small constant $\epsilon_V \ge 0$ such that for all $s, g$:
    \begin{equation}
        |V_\omega(s, g) - V^*(s, g)| \le \epsilon_V
    \end{equation}
    \item \textbf{Bounded anticipation loss:} The anticipation model is trained to a point where its loss, evaluated on the learned value function $V_\omega$, is bounded by a small constant $\epsilon_\psi \ge 0$:
    \begin{equation}
        \max(0, V_\omega(s, g) - V_\omega(s, \hat{s}) - V_\omega(\hat{s}, g) ) \le \epsilon_\psi
    \end{equation}
    where $\hat{s} = \anticipator_\psi(s, g)$.
    \item \textbf{Bounded low-level policy execution:} The low-level policy $\policy_\theta$ is imperfect. When tasked with reaching a subgoal $\hat{s}$ from a state $s$, it induces a distribution over terminal states $s'$. We assume two properties about its execution:
    \begin{enumerate}
        \item The expected cost is bounded by the optimal expected cost plus a sub-optimality gap $\epsilon_\pi \ge 0$:
        \begin{equation}
            \mathbb{E}[C_\theta(s, \hat{s})] \le -V^*(s, \hat{s}) + \epsilon_\pi
        \end{equation}
        \item The stochasticity of the outcome results in a value drift: the expected value of the resulting state $s'$ with respect to the final goal $g$ is close to the value of the intended subgoal $\hat{s}$. We bound this drift by $\epsilon_{drift} \ge 0$:
        \begin{equation}
            |\mathbb{E}_{s' \sim \policy_\theta(\cdot|s,\hat{s})}[V^*(s', g)] - V^*(\hat{s}, g)| \le \epsilon_{drift}
        \end{equation}
    \end{enumerate}
\end{enumerate}
\end{assumption}

This new set of assumptions, particularly 5(b), formally captures the challenge of stochasticity. It states that while the low-level policy may not land on the subgoal, on average, it lands in a region that is equivalently valuable for reaching the final goal, up to a small error $\epsilon_{drift}$. With these more realistic assumptions, we can prove that the agent's expected performance remains bounded.

\begin{theorem}[Bounded Expected Sub-optimality]
\label{thm:main_stochastic}
Under the bounded error conditions for stochastic environments in Assumption \ref{ass:stochastic_error}, the expected total cost $\mathbb{E}[C_{RLA}(s_0, s_g)]$ for the RLA agent to travel from a start state $s_0$ to a goal state $s_g$ over $M$ high-level steps is bounded by:
\begin{equation}
    \mathbb{E}[C_{RLA}(s_0, s_g)] \le -V^*(s_0, s_g) + M \cdot (\epsilon_\pi + 3\epsilon_V + \epsilon_\psi + \epsilon_{drift})
\end{equation}
The term $-V^*(s_0, s_g)$ is the minimum possible expected cost. The theorem states that the agent's expected sub-optimality is bounded by a term that accumulates linearly with the number of high-level planning steps.
\end{theorem}

The proof still relies on the value decomposition property, which holds for stochastic environments.

\begin{lemma}[Value Decomposition]
\label{lemma:bellman}
Let $V^*(s, g)$ be the optimal expected value (negative expected cost) in a stochastic environment. For any three states $s_i, s_j, z$, the value function satisfies the inequality $V^*(s_i, s_j) \ge V^*(s_i, z) + V^*(z, s_j)$. Equality holds if $z$ is an intermediate state on an optimal policy's path from $s_i$ to $s_j$.
\end{lemma}
\begin{proof}
This is a direct consequence of the principle of optimality. Let $\pi^*_{A \to B}$ be an optimal policy for getting from state A to state B, and let $\mathbb{E}[C(\pi)]$ be the expected cost of following policy $\pi$. By definition, $V^*(A, B) = -\mathbb{E}[C(\pi^*_{A \to B})]$. A composite policy that first goes from $s_i$ to $z$ optimally and then from $z$ to $s_j$ optimally has an expected cost of $\mathbb{E}[C(\pi^*_{s_i \to z})] + \mathbb{E}[C(\pi^*_{z \to s_j})]$. Since the optimal policy from $s_i$ to $s_j$ must be at least as good as this composite policy, we have $\mathbb{E}[C(\pi^*_{s_i \to s_j})] \le \mathbb{E}[C(\pi^*_{s_i \to z})] + \mathbb{E}[C(\pi^*_{z \to s_j})]$. Substituting $V^* = -\mathbb{E}[C]$ and multiplying by -1 reverses the inequality, yielding the desired result.
\end{proof}

\begin{proof}[Proof of Theorem \ref{thm:main_stochastic}]
Let the sequence of states where the agent makes high-level decisions be $s_0, s_1, s_2, \dots$. Let $s_k$ be the agent's state at the beginning of the $k$-th high-level step. The agent generates a subgoal $\hat{s}_k = \anticipator_\psi(s_k, s_g)$. It then executes the low-level policy $\policy_\theta(\cdot|s_k, \hat{s}_k)$, which results in a new state $s_{k+1}$, a random variable.

The total expected cost is the sum of the expected costs for each high-level segment:
\begin{equation}
    \mathbb{E}[C_{RLA}(s_0, s_g)] = \mathbb{E}\left[\sum_{k=0}^{M-1} C_k\right]
\end{equation}
where $C_k$ is the cost of the $k$-th segment. Using Assumption \ref{ass:stochastic_error}(5a), we can bound the conditional expectation of each segment's cost:
\begin{equation}
    \mathbb{E}[C_k | s_k] = \mathbb{E}[C_\theta(s_k, \hat{s}_k)] \le -V^*(s_k, \hat{s}_k) + \epsilon_\pi
\end{equation}
We now bound $-V^*(s_k, \hat{s}_k)$. Following the logic of Lemma \ref{lemma:detour} with stochastic values, the bounded anticipation loss (Assumption \ref{ass:stochastic_error}(4)) and bounded value error (Assumption \ref{ass:stochastic_error}(3)) imply:
\begin{equation}
    -V^*(s_k, \hat{s}_k) \le -V^*(s_k, s_g) + V^*(\hat{s}_k, s_g) + 3\epsilon_V + \epsilon_\psi
\end{equation}
Using the value drift assumption (5b) to relate the intended subgoal $\hat{s}_k$ to the actual expected outcome $s_{k+1}$:
\begin{equation}
    V^*(\hat{s}_k, s_g) \le \mathbb{E}[V^*(s_{k+1}, s_g) | s_k] + \epsilon_{drift}
\end{equation}
Substituting this into our cost bound:
\begin{equation}
    -V^*(s_k, \hat{s}_k) \le -V^*(s_k, s_g) + \mathbb{E}[V^*(s_{k+1}, s_g) | s_k] + 3\epsilon_V + \epsilon_\psi + \epsilon_{drift}
\end{equation}
Combining these inequalities and taking the full expectation:
\begin{equation}
    \mathbb{E}[C_{RLA}(s_0, s_g)] \le \mathbb{E} \left[ \sum_{k=0}^{M-1} \left( -V^*(s_k, s_g) + \mathbb{E}[V^*(s_{k+1}, s_g) | s_k] + \epsilon_\pi + 3\epsilon_V + \epsilon_\psi + \epsilon_{drift} \right) \right]
\end{equation}
By the law of total expectation, $\mathbb{E}[\mathbb{E}[V^*(s_{k+1}, s_g) | s_k]] = \mathbb{E}[V^*(s_{k+1}, s_g)]$. The sum $\sum_{k=0}^{M-1} (\mathbb{E}[V^*(s_{k+1}, s_g)] - \mathbb{E}[V^*(s_k, s_g)])$ becomes a telescoping sum in expectation, which evaluates to $\mathbb{E}[V^*(s_M, s_g)] - V^*(s_0, s_g)$. Assuming the process terminates successfully at the goal $s_g$, $\mathbb{E}[V^*(s_M, s_g)] = V^*(s_g, s_g) = 0$. The sum of the value differences thus equals $-V^*(s_0, s_g)$. The total expected cost is therefore bounded by:
\begin{equation}
    \mathbb{E}[C_{RLA}(s_0, s_g)] \le -V^*(s_0, s_g) + M \cdot (\epsilon_\pi + 3\epsilon_V + \epsilon_\psi + \epsilon_{drift})
\end{equation}
This completes the proof.
\end{proof}

\subsection{Discussion}

The primary motivation for RLA's hierarchical decomposition is the potentially dramatic improvement in sample efficiency, a long-standing goal of HRL \cite{bartosweep, pateria2021hierarchical}. 

We can build an intuition of this advantage by drawing on standard results from reinforcement learning theory. The efficiency gain stems from two fundamental properties of stochastic approximation: faster contraction and variance reduction. For faster contraction, the convergence rate of value-based methods is governed by a contraction factor, $\beta$ \cite{bertsekas1996neuro}. In the Stochastic Shortest Path (SSP) setting, this factor is strictly less than 1 but can be arbitrarily close to 1 for problems where the goal is distant, leading to slow convergence \cite{bertsekas1991analysis}. By breaking a long-horizon task (with expected length $L$) into a series of short-horizon sub-tasks (with expected length $K$), RLA ensures that its low-level learner operates on problems with a much better (smaller) contraction factor. The convergence gap $(1-\beta)$ is inversely related to the expected time to reach the goal \cite{szepesvari2010algorithms}, meaning the low-level learner's value function converges at a significantly faster rate. For variance reduction, a flat Q-learner's updates depend on the value of the next state, $V(s', g)$. For a long-horizon task, the variance of this term can be substantial, as the outcome of a single action has a negligible effect on the long-term value. This high variance in the TD target is a well-known factor that slows the convergence of stochastic approximation algorithms \cite{sutton1988learning, kearns1999finite}. In contrast, RLA's low-level learner uses the target $V(s', \hat{s})$, where the subgoal $\hat{s}$ is nearby. Because the sub-task horizon is shorter, the distribution of next-state values has a smaller spread, resulting in a lower-variance learning signal that promotes more stable and efficient learning.

Standard sample complexity bounds for tabular RL formalize this relationship, stating that the number of samples required is proportional to $\sigma^2 / (1-\beta)^2$, where $\sigma^2$ is the variance of the TD-target \cite{azar2013minimax}. Combining these two effects, the sample complexity of RLA's low-level learner is expected to improve over a flat learner by a factor of approximately $(K/L)^2$. This suggests a quadratic gain in sample efficiency from horizon reduction. For a task 100 times longer than the chosen sub-task horizon ($L/K=100$), RLA could be up to 10,000 times more sample-efficient.

This creates a clear and important trade-off for the practitioner. Decreasing the sub-task horizon $K$ yields a powerful, quadratic gain in sample efficiency but incurs a linear penalty in the final policy's asymptotic error bound. The optimal choice of $K$ therefore balances the need for rapid learning against the desire for minimal final sub-optimality, as an overly small $K$ could increase the number of high-level steps $M$ and thus the total accumulated error.

The RLA algorithm detailed in this paper operates in an online learning fashion, where data is generated through continuous exploration. However, the theoretical foundation provided by Theorem \ref{thm:main_stochastic} is not contingent upon this online learning structure. The theorem holds for any learning paradigm, including offline methods, under the condition that the errors in its core assumptions are bounded. This flexibility implies that the subgoal generation mechanism itself is modular. For instance, the anticipation model could be effectively replaced by a planning algorithm that computes optimal subgoals within a given world model, opening the door to model-based hierarchical reinforcement learning approaches.

\section{Related Work}

Our work builds upon several key research areas in reinforcement learning. We situate RLA in the context of hierarchical learning, goal-conditioned policies, and methods that leverage the underlying structure of value functions.

\subsection{Hierarchical reinforcement learning}
RLA is fundamentally an instance of Hierarchical Reinforcement Learning (HRL), a paradigm designed to address long-horizon tasks via temporal abstraction. Foundational work in HRL introduced methods for creating temporally extended actions. The options framework, for example, defines options as policies for specific sub-tasks, complete with their own initiation and termination conditions, allowing them to be treated as primitive actions by a higher-level policy \cite{sutton1999between}. Another seminal approach, MAXQ, decomposes the task's value function into a sum of value functions for constituent sub-tasks, enabling hierarchical value estimation and planning \cite{dietterich2000hierarchical}.

More recent HRL methods have focused on goal-conditioned hierarchies, where a high-level policy directly outputs subgoals in the state space for a low-level policy to achieve. Frameworks like Hierarchical Actor-Critic (HAC) \cite{levy2017hierarchical} and Data-Efficient HRL (HIRO) \cite{nachum2018data} train both levels of the hierarchy using off-policy actor-critic methods. In these architectures, the high-level policy is typically trained via policy gradients, where the reward is the performance of the low-level policy on the assigned subgoal. This approach, however, suffers from non-stationarity, as the high-level policy must learn while the low-level policy's behavior is simultaneously changing. RLA's key distinction is its high-level training objective. The anticipation model is a deterministic function trained on a dense, self-generated signal—geometric consistency with the learned value function—which provides a more stable learning target and is the key to our convergence proof.

\subsection{Temporal abstraction} 
Several recent works formalize or empirically validate long-horizon learning via temporal abstraction and goal-centric representations. Temporal state abstraction and iterated short-horizon supervision improve credit assignment and stability \cite{ghosh2022learning}. On the representation side, time-to-goal/value-as-distance learning and value-function-space abstractions provide geometric inductive biases that align closely with our anticipation objective \cite{hartikainen2020dynamical,shah2021value,eysenbach2022contrastive}. In stochastic shortest-path settings, modern analyses provide near-optimal regret and sample-complexity results \cite{rosenberg2020near,min2022learning,tarbouriech2021sample}, which we leverage to contextualize the efficiency gains of RLA.

\subsection{Hindsight experience replay}
The foundation of our low-level agent is the goal-conditioned policy, a concept popularized by Universal Value Function Approximators (UVFAs), which learn a single policy $\policy(s, g)$ or value function $V(s, g)$ that generalizes over a space of goals \cite{schaul2015universal}. A major breakthrough in training such policies, especially in sparse-reward settings, was Hindsight Experience Replay (HER) \cite{andrychowicz2017hindsight}. The core mechanism of HER is to learn from failure. After an episode where the agent fails to reach its intended goal, HER creates additional hindsight goals from the states the agent actually visited. For a given transition, it relabels the intended goal with a state achieved later in the trajectory, treating that trajectory segment as a success for the new, imagined goal. This creates a dense and effective learning signal from every experience. RLA's low-level module is trained with HER, enabling it to efficiently learn the basic skills required to reach the subgoals proposed by the anticipation model.

\subsection{Unsupervised skill and subgoal discovery}
The anticipation model's role as a subgoal generator connects RLA to the field of intrinsic motivation and automatic subgoal discovery. Many methods in this area learn a repertoire of useful skills without external rewards. Some approaches are driven by curiosity, where an agent builds a model of its environment's dynamics and is intrinsically rewarded for exploring states where its model's predictions are inaccurate \cite{pathak2017curiosity}. Other methods focus on diversity, using information-theoretic objectives to learn a set of skills that are as distinct from one another as possible, thereby encouraging broad exploration of the state space \cite{eysenbach2018diversity}. Unlike these methods, which typically learn a set of general-purpose skills in a task-agnostic manner, RLA's anticipation model is task-directed. It does not generate arbitrary or diverse subgoals; it specifically learns to propose subgoals that are relevant for reaching a given final goal by exploiting the learned value function to find the most efficient path.

\subsection{Learning with value function geometry and successor representations}
Our novel loss function for the anticipation model is inspired by a body of work that exploits the geometric structure of value functions. In environments where value corresponds to negative distance, the optimal value function $V^*(s, g) = -d(s, g)$ must satisfy the triangle inequality. This property has been leveraged in architectures like Value Iteration Networks, which embed a differentiable planning module resembling the value iteration algorithm directly into the policy network \cite{tamar2016value}. Our loss function trains the anticipation model $\anticipator$ to find subgoals $\hat{s}$ that satisfy this geometric condition with equality: $V^*(s_i, s_j) = V^*(s_i, \hat{s}) + V^*(\hat{s}, s_j)$, a condition met if and only if $\hat{s}$ lies on a shortest path.

This concept is also closely related to the Successor Representation (SR) \cite{dayan1993successor, gershman2018successor}. The SR represents a state by the discounted expected future occupancy of all other states. This factorization decouples the environment's dynamics (encoded in the SR) from its rewards, allowing for rapid adaptation when the reward function changes. At its core, the SR encodes reachability information, much like a value function. RLA uses this underlying principle of value-as-distance not just for representation, but as a direct supervisory signal for learning a high-level planner, providing a powerful and theoretically sound alternative to using noisy policy-gradient-based updates.

\section{Conclusion}

In this paper, we introduced Reinforcement Learning with Anticipation (RLA), a hierarchical framework designed to solve complex, long-horizon tasks. RLA's primary contribution is a principled method for training a high-level anticipation model. By training this model to enforce geometric consistency on the agent's own learned value function, we provide a dense and stable learning signal that circumvents many of the instability issues associated with traditional HRL. We further introduced practical mechanisms, such as loss regularization and a training warm-up phase, to handle degenerate solutions and stabilize the challenging joint-training process.

Our central theoretical result is a proof that the RLA system converges to a globally optimal policy under standard, idealized conditions. We also provide bounds on the sub-optimality of the final policy in more realistic settings with function approximation and stochasticity. By decomposing the problem into ``what to do next" (the anticipation model) and ``how to do it" (the low-level policy) and providing a provably correct way to learn both, RLA offers a robust blueprint for creating agents capable of sophisticated, long-range planning.

Future work will focus on extensive empirical studies to validate the practical effectiveness of RLA and its stabilization techniques, and on extending the framework to more complex, high-dimensional environments where these theoretical guarantees can serve as a valuable guide.

\section*{Acknowledgement}

Language and clarity were improved with the assistance of an AI-based writing tool.

\bibliographystyle{plainnat}
\bibliography{ref}

\begin{thebibliography}{28}
\providecommand{\natexlab}[1]{#1}
\providecommand{\url}[1]{\texttt{#1}}
\expandafter\ifx\csname urlstyle\endcsname\relax
  \providecommand{\doi}[1]{doi: #1}\else
  \providecommand{\doi}{doi: \begingroup \urlstyle{rm}\Url}\fi

\bibitem[Andrychowicz et~al.(2017)Andrychowicz, Wolski, Ray, Schneider, Fong,
  Welinder, McGrew, Tobin, Abbeel, and Zaremba]{andrychowicz2017hindsight}
Marcin Andrychowicz, Filip Wolski, Alex Ray, Jonas Schneider, Rachel Fong,
  Peter Welinder, Bob McGrew, Josh Tobin, Pieter Abbeel, and Wojciech Zaremba.
\newblock Hindsight experience replay.
\newblock In \emph{Advances in Neural Information Processing Systems},
  volume~30, 2017.

\bibitem[Azar et~al.(2013)Azar, Munos, and Kappen]{azar2013minimax}
Mohammad~Gheshlaghi Azar, R{'e}mi Munos, and Hilbert~J. Kappen.
\newblock Minimax {PAC} bounds on the sample complexity of reinforcement
  learning with a generative model.
\newblock \emph{Machine Learning}, 91\penalty0 (3):\penalty0 325--349, 2013.

\bibitem[Barto and Mahadevan(2003)]{bartosweep}
Andrew~G. Barto and Sridhar Mahadevan.
\newblock Recent advances in hierarchical reinforcement learning.
\newblock \emph{Discrete Event Dynamic Systems}, 13\penalty0 (4):\penalty0
  341--379, 2003.

\bibitem[Bertsekas and Tsitsiklis(1991)]{bertsekas1991analysis}
Dimitri~P. Bertsekas and John~N. Tsitsiklis.
\newblock An analysis of stochastic shortest path problems.
\newblock \emph{Mathematics of Operations Research}, 16\penalty0 (3):\penalty0
  580--595, 1991.

\bibitem[Bertsekas and Tsitsiklis(1996)]{bertsekas1996neuro}
Dimitri~P. Bertsekas and John~N. Tsitsiklis.
\newblock \emph{Neuro-Dynamic Programming}.
\newblock Athena Scientific, 1996.

\bibitem[Cohen et~al.(2020)Cohen, Kaplan, Mansour, and
  Rosenberg]{rosenberg2020near}
Alon Cohen, Haim Kaplan, Yishay Mansour, and Aviv Rosenberg.
\newblock Near-optimal regret bounds for stochastic shortest path.
\newblock In \emph{Advances in Neural Information Processing Systems},
  volume~33, 2020.

\bibitem[Dayan(1993)]{dayan1993successor}
Peter Dayan.
\newblock Improving generalization for temporal difference learning: The
  successor representation.
\newblock \emph{Neural Computation}, 5\penalty0 (4):\penalty0 613--624, 1993.

\bibitem[Dietterich(2000)]{dietterich2000hierarchical}
Thomas~G. Dietterich.
\newblock Hierarchical reinforcement learning with the {MAXQ} value function
  decomposition.
\newblock \emph{Journal of Artificial Intelligence Research}, 13:\penalty0
  227--303, 2000.

\bibitem[Eysenbach et~al.(2019)Eysenbach, Gupta, Ibarz, and
  Levine]{eysenbach2018diversity}
Benjamin Eysenbach, Abhishek Gupta, Julian Ibarz, and Sergey Levine.
\newblock Diversity is all you need: Learning skills without a reward function.
\newblock In \emph{Proceedings of the 7th International Conference on Learning
  Representations}, 2019.

\bibitem[Eysenbach et~al.(2022)Eysenbach, Zhang, Salakhutdinov, and
  Levine]{eysenbach2022contrastive}
Benjamin Eysenbach, Tianjun Zhang, Ruslan Salakhutdinov, and Sergey Levine.
\newblock Contrastive learning as goal-conditioned reinforcement learning.
\newblock In \emph{Advances in Neural Information Processing Systems},
  volume~35, 2022.

\bibitem[Gershman(2018)]{gershman2018successor}
Samuel~J. Gershman.
\newblock The successor representation: Its computational logic and neural
  substrates.
\newblock \emph{Journal of Neuroscience}, 38\penalty0 (33):\penalty0
  7193--7200, 2018.

\bibitem[Ghosh et~al.(2019)Ghosh, Gupta, Reddy, Fu, Devin, Eysenbach, and
  Levine]{ghosh2022learning}
Dibya Ghosh, Abhishek Gupta, Ashwin Reddy, Justin Fu, Coline Devin, Benjamin
  Eysenbach, and Sergey Levine.
\newblock Learning to reach goals via iterated supervised learning.
\newblock arXiv preprint arXiv:1912.06088, 2019.

\bibitem[Hartikainen et~al.(2020)Hartikainen, Geng, Haarnoja, and
  Levine]{hartikainen2020dynamical}
Kristian Hartikainen, Xinyang Geng, Tuomas Haarnoja, and Sergey Levine.
\newblock Dynamical distance learning for semi-parametric control.
\newblock In \emph{Proceedings of the 8th International Conference on Learning
  Representations}, 2020.

\bibitem[Kearns and Singh(1999)]{kearns1999finite}
Michael Kearns and Satinder Singh.
\newblock Finite-sample convergence rates for {Q}-learning.
\newblock In \emph{Advances in Neural Information Processing Systems},
  volume~11, pages 996--1002, 1999.

\bibitem[Levy et~al.(2019)Levy, Konidaris, Platt, and
  Saenko]{levy2017hierarchical}
Andrew Levy, George Konidaris, Robert Platt, and Kate Saenko.
\newblock Learning multi-level hierarchies with hindsight.
\newblock In \emph{Proceedings of the 7th International Conference on Learning
  Representations}, 2019.

\bibitem[Lillicrap et~al.(2016)Lillicrap, Hunt, Pritzel, Heess, Erez, Tassa,
  Silver, and Wierstra]{lillicrap2015continuous}
Timothy~P. Lillicrap, Jonathan~J. Hunt, Alexander Pritzel, Nicolas Heess, Tom
  Erez, Yuval Tassa, David Silver, and Daan Wierstra.
\newblock Continuous control with deep reinforcement learning.
\newblock In \emph{Proceedings of the 4th International Conference on Learning
  Representations}, 2016.

\bibitem[Min et~al.(2022)Min, He, Wang, and Gu]{min2022learning}
Yifei Min, Jiafan He, Tianhao Wang, and Quanquan Gu.
\newblock Learning stochastic shortest path with linear function approximation.
\newblock In \emph{Proceedings of the 39th International Conference on Machine
  Learning}, volume 162 of \emph{Proceedings of Machine Learning Research},
  pages 15608--15639, Baltimore, MD, 2022. PMLR.

\bibitem[Nachum et~al.(2018)Nachum, Gu, Lee, and Levine]{nachum2018data}
Ofir Nachum, Shixiang Gu, Honglak Lee, and Sergey Levine.
\newblock Data-efficient hierarchical reinforcement learning.
\newblock In \emph{Advances in Neural Information Processing Systems},
  volume~31, 2018.

\bibitem[Pateria et~al.(2021)Pateria, Subagdja, Tan, and
  Quek]{pateria2021hierarchical}
Shubham Pateria, Budhitama Subagdja, Ah-Hwee Tan, and Chai Quek.
\newblock Hierarchical reinforcement learning: A comprehensive survey.
\newblock \emph{ACM Computing Surveys}, 54\penalty0 (5):\penalty0 1--35, 2021.

\bibitem[Pathak et~al.(2017)Pathak, Agrawal, Efros, and
  Darrell]{pathak2017curiosity}
Deepak Pathak, Pulkit Agrawal, Alexei~A. Efros, and Trevor Darrell.
\newblock Curiosity-driven exploration by self-supervised prediction.
\newblock In \emph{Proceedings of the 34th International Conference on Machine
  Learning}, volume~70 of \emph{Proceedings of Machine Learning Research},
  pages 2778--2787. PMLR, 2017.

\bibitem[Schaul et~al.(2015)Schaul, Horgan, Gregor, and
  Silver]{schaul2015universal}
Tom Schaul, Daniel Horgan, Karol Gregor, and David Silver.
\newblock Universal value function approximators.
\newblock In \emph{Proceedings of the 32nd International Conference on Machine
  Learning}, volume~37 of \emph{Proceedings of Machine Learning Research},
  pages 1312--1320. PMLR, 2015.

\bibitem[Shah et~al.(2022)Shah, Xu, Lu, Xiao, Toshev, Levine, and
  Ichter]{shah2021value}
Dhruv Shah, Peng Xu, Yao Lu, Ted Xiao, Alexander~T. Toshev, Sergey Levine, and
  Brian Ichter.
\newblock Value function spaces: Skill-centric state abstractions for
  long-horizon reinforcement learning.
\newblock In \emph{Proceedings of the 10th International Conference on Learning
  Representations}, 2022.

\bibitem[Sutton(1988)]{sutton1988learning}
Richard~S. Sutton.
\newblock Learning to predict by the methods of temporal differences.
\newblock \emph{Machine Learning}, 3\penalty0 (1):\penalty0 9--44, 1988.

\bibitem[Sutton et~al.(1999)Sutton, Precup, and Singh]{sutton1999between}
Richard~S. Sutton, Doina Precup, and Satinder Singh.
\newblock Between {MDP}s and semi-{MDP}s: A framework for temporal abstraction
  in reinforcement learning.
\newblock \emph{Artificial Intelligence}, 112\penalty0 (1--2):\penalty0
  181--211, 1999.

\bibitem[Szepesv{'a}ri(2010)]{szepesvari2010algorithms}
Csaba Szepesv{'a}ri.
\newblock \emph{Algorithms for Reinforcement Learning}.
\newblock Morgan \& Claypool Publishers, 2010.

\bibitem[Tamar et~al.(2016)Tamar, Wu, Thomas, Levine, and
  Abbeel]{tamar2016value}
Aviv Tamar, Yi~Wu, Garrett Thomas, Sergey Levine, and Pieter Abbeel.
\newblock Value iteration networks.
\newblock In \emph{Advances in Neural Information Processing Systems},
  volume~29, 2016.

\bibitem[Tarbouriech et~al.(2021)Tarbouriech, Pirotta, Valko, and
  Lazaric]{tarbouriech2021sample}
Jean Tarbouriech, Matteo Pirotta, Michal Valko, and Alessandro Lazaric.
\newblock Sample complexity bounds for stochastic shortest path with a
  generative model.
\newblock In \emph{Proceedings of the 32nd International Conference on
  Algorithmic Learning Theory}, volume 132 of \emph{Proceedings of Machine
  Learning Research}, pages 1--22. PMLR, 2021.

\bibitem[Watkins and Dayan(1992)]{watkins1992q}
Christopher J. C.~H. Watkins and Peter Dayan.
\newblock {Q}-learning.
\newblock \emph{Machine Learning}, 8\penalty0 (3):\penalty0 279--292, 1992.

\end{thebibliography}

\end{document}